\documentclass[lettersize,journal]{IEEEtran}
\usepackage{amsmath,amsfonts}
\usepackage{algorithmic}
\usepackage{algorithm}
\usepackage{array}
\usepackage[caption=false,font=normalsize,labelfont=sf,textfont=sf]{subfig}
\usepackage{textcomp}
\usepackage{stfloats}
\usepackage{url}
\usepackage{verbatim}
\usepackage{graphicx}
\usepackage{cite}
\begin{document}

\title{A Graph Neural Architecture Search Approach for Identifying Bots in Social Media}
\author{Georgios Tzoumanekas, Michail Chatzianastasis, Loukas Ilias, George Kiokes, John Psarras, Dimitris Askounis
\thanks{G. Tzoumanekas, L. Ilias, J. Psarras, and D. Askounis are with the Decision Support Systems Laboratory, School of Electrical and Computer Engineering, National Technical University of Athens, 15780 Athens, Greece (e-mail: lilias@epu.ntua.gr; askous@epu.ntua.gr; john@epu.ntua.gr). 

M. Chatzianastasis is with DaSciM, LIX, Ecole Polytechnique, Institut Polytechnique de Paris, Palaiseau, 91120, France (email: mixalisx97@gmail.com). 

G. Kiokes is with the Laboratory of Electrical Machines and Installations, Division of Electrical, Electronics and Informatics, School of Engineering, Merchant Marine Academy of Aspropyrgos, 19300 Aspropyrgos, Greece (email: gkiokes@iccs.gr).}}



\maketitle

\begin{abstract}
Social media platforms, including X, Facebook, and Instagram, host millions of daily users, giving rise to bots-automated programs disseminating misinformation and ideologies with tangible real-world consequences. While bot detection in platform X has been the area of many deep learning models with adequate results, most approaches neglect the graph structure of social media relationships and often rely on hand-engineered architectures. Our work introduces the implementation of a Neural Architecture Search (NAS) technique, namely Deep and Flexible Graph Neural Architecture Search (DFG-NAS), tailored to Relational Graph Convolutional Neural Networks (RGCNs) in the task of bot detection in platform X. Our model constructs a graph that incorporates both the user relationships and their metadata. Then, DFG-NAS is adapted to automatically search for the optimal configuration of Propagation and Transformation functions in the RGCNs. Our experiments are conducted on the TwiBot-20 dataset, constructing a graph with 229,580 nodes and 227,979 edges. We study the five architectures with the highest performance during the search and achieve an accuracy of 85.7\%, surpassing state-of-the-art models.
Our approach not only addresses the bot detection challenge but also advocates for the broader implementation of NAS models in neural network design automation.
\end{abstract}


\section{Introduction}
Social media are online community platforms and apps that let users create, share, and interact with each other's content. Social media content can be text, photos, videos, GIFs, audio, etc. Social media can be used for various reasons, from users who share interests communicating to getting informed about current worldwide events. Social media can also be used for detecting early signs of stress and depression \cite{ILIAS2023100270,10154134, kerasiotis2024}. The existence of social media in our day-to-day lives is more prevalent than ever. As of 2023, there are roughly 4.9 billion social media users, a percentage that is more than 60\% of the entire population and more than 100 social media platforms. X, previously known as Twitter, stands out as one of the most widely recognized social media platforms. Twitter was launched in 2006. It revolves around the concept of “following” other users. A user can follow accounts they are interested in and see their tweets in their timeline (“following”) and conversely can have “followers” that see their tweets. Nowadays, it has been established as a powerful tool for real-time news updates, public discourse, and social movements, and continues to evolve and enhance its user experience. In 2023, Twitter was renamed to X by then-CEO Elon Musk. The extensive presence of social media in the modern landscape has led to the emergence of accounts that automate interactions on social media platforms, often mimicking human behavior, the so-called bots. These bots can be coded to perform a variety of tasks, such as automatically publishing content, liking, sharing, following, or commenting on posts. Some can even be programmed to engage in conversations to promote specific agendas.
 Their behavior differs depending on their intent and purpose, but they might share features, such as very high or very low activity levels and more structured and characteristic language patterns \cite{introrussian}. Uyheng et al. \cite{introtrolling} examined the origin and traits of trolling messages, finding that they often originate from automated bots and are distinguished by their use of abusive language, reduced cognitive complexity, and specific targeting of individuals or entities. Their study also noted a tendency for bots to target right-leaning sources of information, while trolls tended to engage with less polarized content, spreading misinformation across diverse audiences. Bots are very efficient in spreading misinformation, particularly when programmed with optimized values for factors like walking speed, network distribution, and strategy \cite{introstrategy}.
Fake news and bots have had significant tangible consequences in several cases.  Users tend to believe conspiracy theories and misinformation, and correction attempts can sometimes backfire \cite{introgatekeeping}. Users might also share fake news for altruistic or self-promotional purposes, yet those with greater social media literacy are better equipped to identify and refrain from spreading fake news \cite{introknowledge}. Therefore, there's a need for measures to promote truthful reporting in media and detect any cases of misinformation dissemination.

The need to detect bot accounts to shut them down is quite immediate, assessing the hazards of their uncontrollable presence on social media. Several studies to identify bots from real users have been conducted that provide satisfactory results. There have been several approaches, including supervised learning \cite{relatedlee}, unsupervised learning \cite{relatedcresci}, reinforcement learning \cite{relatedalauthman}, and GNN-based architectures \cite{botrgcn}. However, all these traditional neural architectures often rely on fixed parameters that are manually designed. Constructing efficient neural network architectures requires extensive feature engineering and can be a quite challenging and time-consuming procedure. Also, fixed architectures often mitigate the models' adaptability on other datasets and tasks. Motivated by these limitations, we examine the implementation of Neural Architecture Search (NAS) to automate the process of discovering optimal architectures. NAS explores a search space of possible architectures and identifies the configurations that enhance the model's performance.

A Neural Architecture Search method that has been proposed to solve the performance issues of fixed architectures is Deep and Flexible Graph Neural Architecture Search (DFG-NAS) \cite{dfgnas}. It employs an evolutionary algorithm to explore a vast space of permutations of Propagation and Transformation operations, to find the one with the best accuracy in the validation set. Addressing the limitations of previous bot detection models due to their fixed architectures we employ DFG-NAS on a GNN-based approach for bot detection. This approach leverages the user's semantical and property information and constructs a heterogeneous graph out of the follower-following relationships between users. Then, we adapt the DFG-NAS approach to handle Relational Graph Convolutional Neural Networks (RGCNs). The model automatically searches for the permutation of Propagation (P) and Transformation (T) functions, the two main processes of the message-passing protocol, with the highest validation accuracy. The model is also amplified with the use of the Gate operation on the P connections and the use of the skip-connection operation on the T connections.

To the authors' knowledge, DFG-NAS has not been employed before in the task of bot detection. All our experiments were performed on the Twibot-20 dataset \cite{twibotdataset}. The following sums up the contributions of our work:

\begin{itemize}
    \item We implement DFG-NAS, tailored to RGCNs, to automatically determine the most effective permutation of the message-passing operations.
    \item We perform experiments to demonstrate the benefits of architecture search in bot detection and compare our method to state-of-the-art models.
    \item We perform a thorough ablation study on the necessity of the user metadata in our graph, the Gate operation, and the skip-connection operation in NAS.
\end{itemize}

\section{Research objective}

It is evident to any social media user that bots continue to dominate the digital landscape despite extensive efforts in bot detection and platform initiatives to stop their activities. As technology advances, bots are programmed to mimic human mannerisms more effectively, making them more resilient against detection mechanisms. Beyond the irritation they pose to everyday users, some bots can have tangible and detrimental effects on human society. In 2016 fake news stories spread widely during the U.S. presidential election campaign aiming to influence the public vote \cite{intro1}. Throughout the COVID-19 pandemic \cite{intro2}, bots spread misinformation about the virus and the vaccines on social media, leading to mob panic, confusion, and even resistance to public health measures.  Fake news is often framed in a manner that fosters negativity in social discussions and hinders individuals' ability to consider diverse perspectives, contributing to the formation of 'echo chambers' on social media platforms \cite{introecho}. Bots also exacerbate cyberbullying by mass-targeting users, leading to serious psychological consequences. Social media platforms face challenges in effectively moderating such content. Cyberbullying detection methods often rely on unclear definitions and are prone to biases in data annotation \cite{introcyber}. Their evolving nature raises concerns about the efficiency of current preventive measures, highlighting the need for innovation to prevent the dangers posed by this digital phenomenon.

The motivation for this research was constructing a model characterized by adaptability across future datasets, ensuring resilience in the face of evolving technology through time. Many contemporary models rely on fixed architectures, often struggling to demonstrate their efficiency on novel datasets. Although Neural Architecture Search (NAS) has shown significant advantages in various test cases, its application to bot detection remains relatively underexplored, with limited but promising results noted in studies such as \cite{rosgas}. Considering the dynamic nature of the social media landscape and the continuous evolution of bots, more flexible architectures specifically designed for bot identification could offer a practical solution to mitigate their real-world consequences.

This research aims to showcase the efficiency advantages of architecture search and perhaps pave the way for more implementations of NAS models in bot detection in the ongoing battle against automated malicious activities.

\section{Related Work}
\subsection{Bot and fake news detection models}
The task of bot identification has attracted numerous studies and many state-of-the-art models propose fascinating methodologies. We could mainly divide these models into supervised learning approaches, unsupervised learning approaches, and GNN-based approaches. In this section, we present some baseline models proposed for bot detection and discuss how they fall into the above categories.

Lee et al. \cite{relatedlee} applied various machine learning algorithms, including SVMs, Naive Bayes, and decision trees, to build and evaluate a supervised bot detection model. The features used in their analysis included account-based features (e.g., the number of followers, friends, tweets), temporal features (e.g., time of account creation, tweet frequency), and content-based features (e.g., usage of URLs, hashtags). Kuduganta et al. \cite{relatedkuduganta} suggested a deep learning model that uses the user's tweets and some metadata features. This architecture includes a tokenizer, GloVE embedding layer, LSTM, and Dense layers. Wei et al. \cite{relatedwei} used only users' tweets with no context of prior knowledge on user profiles, friendship networks, or behaviour. They proposed a recurrent neural network (RNN) model that used word embeddings to encode tweets, a three-layer Bidirectional LSTM (BiLSTM), and a softmax layer at the binary output. Cai et al. \cite{relatedcai} proposed their model (BeDM) that involved deep neural networks in bot detection. They employed convolutional neural networks (CNNs) and LSTM, using only the tweet semantics, such as the frequency and the type of publications. Botometer \cite{relatedbotometer} is a web-based program developed by Davis et al. at Indiana University. It leverages more than 1,000 features to classify Twitter accounts as bots and humans, such as friends, the structure of the social network, user meta-data, temporal activity, and sentiment analysis. Botometer distinguishes the accounts by an overall bot score (ranging from 0 to 5), along with several other scores. The greater the score, the greater the probability that this account is linked to a bot.  Yang et al. \cite{yang2022botometer} presented a thorough introduction of the latest version of Botometer for new users and demonstrated a case study. Alarfaj et al. \cite{relatedalarfaj} utilized features based on content attained from the Twitter API and employed state-of-the-art classifiers, like MLPs, random forest, and naive Bayes. Features included messages, special characters, sentiment analysis, etc. Alothali et al. \cite{relatedalothali} introduced their framework, called Bot-MGAT, which stands for bot multi-view graph attention network. The scientists pointed out that other approaches couldn't adjust to different datasets since there wasn't enough recently updated labeled data, which made sense given the constantly shifting behavior of the bots. They presented a methodology that makes use of transfer learning (TL) to leverage the multi-view graph attention mechanism. The framework also benefited from semi-supervised learning, using labeled and unlabeled data. The authors used the TwiBot-20 \cite{twibotdataset} due to its graph structure, extracting 18 features for the training. Feng et al. \cite{relatedsatar} suggested SATAR. In particular, SATAR leverages the user's semantics, property, and neighborhood information. It adjusts by fine-tuning parameters and pre-training on a huge number of self-supervised users. The authors proposed two alternative models: $SATAR_{FC}$ and $SATAR_{FT}$. Ilias et al. \cite{relatedilias2} proposed two methods for bot detection using deep learning techniques. Their first approach extracts a substantial 71 features per user to utilize for account classification to bots and genuine users. They also employed various feature selection techniques to discard redundant and irrelevant features. Their second methodology proposes a deep learning architecture for tweet-level classification. This architecture incorporates an attention mechanism atop the Bidirectional Long Short-Term Memory (BiLSTM) layer. During the learning phase, the attention mechanism helps the model better focus on relevant information. Ilias et al. \cite{relatedilias} focused solely on user descriptions and sequences of actions performed by Twitter accounts. Their approach includes both unimodal (text or image) and multimodal (both text and image) methods. They designed digital DNA sequences per user based on tweet type and content, converted these sequences into 3D images, and fine-tuned pre-trained vision models like AlexNet, ResNet, and VGG16. For bot detection through user descriptions, they fine-tuned TwHIN-BERT, a transformer model. In multimodal approaches, they use VGG16 for visual representation and TwHIN-BERT for textual representation, proposing three fusion methods: concatenation, gated multimodal unit (GMU), and cross-attention. They conducted their experiments on the Cresci’17 dataset. Wei et al. \cite{relatedweinguyen} proposed their model BOTLE. Their model utilizes a recurrent neural network (RNN) with Bidirectional Gated Recurrent Units (BiLGRU) connecting two hidden layers of opposite directions leading to the same output. Notably, BOTLE does not rely on handcrafted features or pre-existing information regarding account profiles. Linguistic embeddings, including word, character, part-of-speech, and named-entity embeddings, are employed to encode tweet content, eliminating the need for labor-intensive feature engineering.  Bazmi et al. \cite{relatedmvcan} introduced the Multi-View Co-Attention Network (MVCAN), which aims to capture the latent topic-specific credibility of both users and news sources. This model represents news articles, users, and news sources in a manner that encodes topical viewpoints, socio-cognitive biases, and partisan biases as vectors. These features are encoded using a variant of the Multi-Head Co-Attention (MHCA) mechanism. Shevtsov et al. \cite{relatedbotartist} introduced their model BotArtist, constructed on a semi-automatic machine learning pipeline, that requires minimal features for training, taking into consideration the loads of data needed by previous approaches and the recent monetization of Twitter API requests. Sujith et al. \cite{relatedsujith} proposed a supervised learning approach that used multiple models to detect bots. Their classification of accounts relied on features like user metadata, tweet content, and posting history, among others. In addition to identifying bot accounts, the authors assigned a level of significance or influence to them, prioritizing the removal of the most influential or harmful bot accounts. Liu et al. \cite{relatedliu} proposed BotMoE, which leverages three perspectives of user information (metadata, text, and graph representations) and incorporates a community-aware Mixture-of-Experts (MoE) layer to assign users to different communities. The user representations are fused with an extractor fusion layer and supervised learning is employed to train the BotMoE framework to perform community-aware bot detection. Saxena et al. \cite{relatedsaxena} proposed two frameworks for recognizing accounts that disseminate false information on Twitter. Initially, they employed profile-based data, including the verified status, profile photo, and account lifetime and activity. Then, they combined tweet-propagation patterns and assigned a credibility score to each user, signifying their authenticity. Dimitriadis et al. \cite{relatedcaleb} proposed CALEB that is based on the Conditional Generative Adversarial Network (CGAN) and its extension, Auxiliary Classifier GAN (AC-GAN). By developing realistic artificial bot varieties, they were able to replicate the evolution of bots. As a result, they enhanced already-existing datasets and were able to identify bots before they emerged.

Yang et al. \cite{relatedyang} used a combination of unsupervised and supervised learning methods for bot detection. Specifically, the authors utilized minimal features derived from user metadata, temporal patterns, network structure, sentiment analysis, and linguistic cues that they fed into a machine learning pipeline, that reduced dimensionality and included classification algorithms. Cresci et al. \cite{relatedcresci} introduced the Social Fingerprinting technique for bot detection, a Digital DNA technique that models social network users' behaviors. Each user is represented as a sequence of characters depending on the type and content of the tweets they publish, simulating a DNA sequence. The authors try to find similarities in the sequences defining the length of the Longest Common Substring (LCS) between two sequences. For a set of real users, the length of LCS was found to be particularly small, leading to the conclusion that longer sequences than the average LCS were bots. Based on this idea, the authors developed two techniques, one based on supervised learning and another on unsupervised learning to find similarities in the behaviour of accounts.  Quezada et al. \cite{relatedquezada} developed a real-time bot infection detection model that analyzes Domain Name System (DNS) traffic events. They extracted 13 attributes from DNS logs to create unique fingerprints for servers. Using Isolation Forest, an algorithm for unsupervised learning, they identified anomalies in the fingerprints to classify hosts as infected or not. The model also utilized Domain Generation Algorithms (DGA) to search for queries to anomalous domains. Finally, a Random Forest, a supervised learning algorithm, was employed to create a model for detecting future bot infections on hosts.
Miller et al. \cite{relatedmiller} approached bot identification as an anomaly detection problem. They extracted 107 features from user’s tweets and property information and adapted two stream clustering algorithms, StreamKM++ and DenStream, to facilitate spam detection and identified bot users as abnormal outliers. Chavoshi et al. \cite{relatedchavoshi} developed DeBot, a bot detection system for social media, using warped correlation to identify likely bot accounts based on their high synchronicity, a characteristic unlikely in human users. DeBot doesn't require labeled data and operates on activity correlation. Moreover, through the utilization of a lag-sensitive hashing technique, it can promptly cluster accounts for real-time classification.  Minnich et al. \cite{relatedbotwalk} proposed their real-time unsupervised model BotWalk. Using metadata, content, temporal, and network features they employ anomaly detection, comparing each user to a seed bank of labeled accounts iteratively. Mannocci et al. \cite{relatedmulbot} proposed MulBot, an unsupervised bot detection system that utilizes multivariate time series (MTS) analysis. They employed an LSTM autoencoder to map the MTS data into a latent space and then conducted clustering on this encoded data to find dense clusters of users exhibiting similar behavior, assuming this was a common trait of bot accounts. MulBot also showcases effectiveness in identifying and distinguishing various botnets. Wu et al. \cite{relatedwu} employed unsupervised machine learning techniques, specifically K-Means and Agglomerative clustering, for Twitter bot detection. They used account activity, popularity, and verification status, among other features for the clustering. Koggalahewa et al. \cite{relatedkoggalahewa} introduced an unsupervised method for bot identification based on a user's peer approval in the social network. They based peer acceptance between two users on their shared interests over a multitude of issues. Lopes et al. \cite{relatedlopes} introduced their botnet identification model, designed to detect networks of compromised devices under master control. Their approach relies on analyzing network flow behavior through a contemporary method known as the Energy-based Flow Classifier (EFC). EFC employs inverse statistics to enhance anomaly detection.

 Alhosseini et al. \cite{relatedalhosseini} introduced the use of graph convolutional neural networks (GCNN) in bot identification. They noted that besides the users' features, the construction of a social network would enhance a model's ability to distinguish the bots from the genuine users. Feng et al. \cite{relatedheterofeng} introduced the aspect of diversity in relationships and influence dynamics among users in the Twittersphere for bot detection. They proposed a bot detection framework that leverages a network with users as nodes and the different relations as edges. Then they aggregated messages across users and operated heterogeneity-aware Twitter bot detection. They conducted their experiments using the Twi-Bot20 dataset. Feng et al. \cite{botrgcn} proposed their model for bot detection BotRGCN, which is short for Bot detection with Relational Graph Convolutional Networks. BotRGCN builds a heterogeneous graph out of the following relationships and uses information, such as the user's description, tweets, numerical and categorical property set, and neighborhood information. The experiments were conducted on the Twi-Bot20 dataset \cite{twibotdataset}, but BotRGCN could exploit other types of relations if supported by the dataset.  Kušen et al. \cite{relatedkusen} examined the structural dynamics of conversations between humans and bots on Twitter following emotionally charged riot events. They introduced "emotion-exchange motifs" to identify recurring patterns in emotional message exchanges. Their findings revealed that human conversations exhibited various motifs with reciprocal edges and self-loops, indicating interactive dialogue. In contrast, bots typically disseminated identical messages to multiple users or did not anticipate replies. Moreover, bots frequently initiated conversations and often conveyed fear-inducing messages.
Bui et al. \cite{relatedbui} introduced a graph-based method for bot detection. They detailed their data collection process and identified specific behaviors indicative of an account being associated with a bot. These behaviors can include engagement with other users, nonsensical usernames and profile information, repetitive content posting, and retweeting activity. These observations are utilized to label the accounts accordingly. Dehghan et al. \cite{relateddehghan} suggested that the local social network formed around each account can aid in identifying the bots. To prove their hypothesis, they compared two classes of embedding algorithms, the former of which focused on proximity data and the latter that focused on nodes' neighborhoods. They discovered that the structural embeddings presented higher information underlining the valuable information that is embedded within each node's local network. Pham et al. \cite{relatedbot2vec} introduced their approach Bot2Vec, which eliminated the need for user profile features. To improve the model's generalization on many social media platforms, they used only local neighborhood relations and the community structure of the graph that represented the users and employed an random walk strategy in the communities. Noekhah et al. \cite{relatednoekhah} proposed their model "Multi-iterative Graph-based opinion Spam Detection" (MGSD) that aims to identify various types of spam entities. It analyzes all kinds of relationships between them and utilizes domain-independent features, allowing for generalization across types of opinionated documents. Trained on both existing and novel features, MGSD assigns a spam score to each entity. Ye et al. \cite{relatedhofa} proposed HOFA, a graph-based framework for bot detection, featuring two key modules: Homophily-Oriented Graph Augmentation (Homo-Aug) and Frequency Adaptive Attention (FaAt). The Homo-Aug employs an MLP to extract user representations and generate a k-NN graph. Meanwhile, the FaAt module acts as a low-pass filter for homophilic edges and a high-pass filter for heterophilic edges. This function prevents excessive smoothing of user features by the neighborhood. El-Mawass et al. \cite{relatedsimilcatch} explored using the output of existing supervised classification systems to detect spammers. They incorporated the classifiers' outputs as prior beliefs within a probabilistic graphical model framework. Proposing a bipartite users-content interaction graph, they facilitated the spread of beliefs to similar accounts. Constructing a Markov Random Field on a graph of similar users, they employed Loopy Belief Propagation to derive the predictions. Their findings demonstrated a notable enhancement in recall while maintaining precision.

\subsection{Neural Architecture Search approaches}
Graph neural architecture search is proposed as the solution to performance limitations due to a fixed architecture. Parameter tuning in neural networks can be a challenging task. Many NAS methods have been suggested that include variations in the search space, the optimization method, and the architecture evaluation. We will divide these methods based on their optimization method, which will include reinforcement learning, evolutionary algorithms and gradient-based methods.

Zhou et al. \cite{relatedzhou} proposed the automated graph neural networks (Auto-GNN) framework. Auto-GNN searches for the best GNN architecture possible in a predetermined search space, divided into six classes of actions: hidden dimension, attention function, attention head, aggregate function, combine function, and activation function. For efficiency reasons, the authors designed a conservative explorer to preserve the optimal neural architecture discovered during the search. The authors also implemented constrained parameter sharing, adapted to the heterogeneous GNN architecture. Two experimental methods were presented: inductive, in which the graph structure and node features on the testing and validation sets are unknown during training, and transductive, which involves the availability of unlabeled data for testing and validation during training. Gao et al. \cite{relatedgraphnas} proposed GraphNAS to implement an automatic search of the best graph neural architecture based on reinforcement learning. The search space covers sampling functions, aggregation functions, and gated functions. GraphNas also uses more efficient parameter-sharing techniques than other contiguous models for CNNs and RNNs. After training 1000 different architectures, the five best ones were used for the testing, which surpassed human-invented ones or those produced by random searches. Zhao et al. \cite{relatedsnag} proposed the SNAG framework (Simplified Neural Architecture Search for Graph neural networks). The suggested framework had two key components: Node aggregators, which focused on neighborhood features, and Layer aggregators, which focused on the range of the neighborhood used. The search space algorithm was a variant of Reinforcement Learning that adopted the weight-sharing mechanism (SNAGWS). Nunes et al. \cite{relatednunes} presented one NAS methods for optimizing GNNs based on reinforcement learning and one based on evolutionary algorithms. The authors defined two cases of search spaces: Macro, where the architectures generated have the same structure, and Micro, where the architectures are not rigidly structured but combine several convolutional schemas. They concluded that EA and RL found very similar architectures to those found by a random search, a significantly simpler technique. However, they pointed out that whilst the other approaches generated large structures in as much as 80\% of the situations, EA created the majority of GPU-fitting designs.  Li et al. \cite{relatedmetagnas} proposed Meta-GNAS that uses meta-reinforcement learning from past tasks to apply that knowledge to new tasks. Additionally, they speed up the search by using a predictive model to evaluate the potential graph neural architectures instead of training them from scratch.

Peng et al. \cite{relatedpeng} implemented a NAS approach to human action recognition from skeleton movements. The search space was enlarged with diverse spatial-temporal graph modules while constructing higher-order connections between nodes using Chebyshev polynomial approximation. The search algorithm used is an evolutionary adaptation with a high sampling efficiency, denoted CEIM (Cross-Entropy method with ImportanceMixing). Jiang et al. \cite{relatedjiang} adapted the method of neural architecture search to the conception of GNNs for predicting molecular properties. The authors designed neural networks for message-passing (MPNNs) between nodes. To find an optimal MPNN from the user-defined search space, they used regularized evolution (RE) from the DeepHyper package. Zhang et al. \cite{dfgnas} proposed DFG-NAS, an innovative method that allows for automatic search of very deep and adaptable GNN architectures. DFG-NAS focuses on exploring macro-architectures, specifically the implementation details of atomic propagation (P) and transformation (T) operations within the GNN. P is linked to the graph structure, whereas T concentrates on the non-linear transformations within the neural network. In addition, they adopted gating and skip-connection mechanisms for deeper GNN pipelines. They used an evolutionary algorithm to find the optimal architecture, which supported four cases of mutation.  Peng et al. \cite{relatedfastenas} introduced Fast-ENAS as a computationally efficient alternative to Evolutionary Neural Architecture Search. This method utilizes a training-free performance metric that is computed with a single forward pass. The authors enhance the search process by incorporating a GCN-based contrastive predictor, aiming to improve the accuracy of the estimated performance of a candidate architecture, bringing it closer to its actual performance. Shang et al. \cite{relatedshang} introduced AG-ENAS, which brings two key innovations to the Evolutionary Neural Architecture Search process. Firstly, it employs an adaptive parameter adjustment mechanism based on population diversity and fitness, enhancing the adaptation of genetic operators' associated parameters. Secondly, the model introduces a mutation operator guided by the gene potential contribution. It improves offspring quality by assigning weight to more valuable genes through a distribution index matrix. The concept of aging is integrated into environmental selection to mitigate premature convergence. Lopes et al. \cite{relatedgea} presented GEA (Guided Evolutionary Architecture), which tackles the problem of other NAS models getting trapped in suboptimal solutions during the search process. GEA overcomes this challenge by generating and evaluating multiple architectures using a zero-proxy estimator and selecting only one with the best-performing one for the next generation. This approach expands the search space without increasing complexity, as new architectures are derived from previous ones through mutations.

Zhao et al. \cite{relatedsane} proposed their framework SANE. The search space has similarities with the search space from the SNAG framework, with Node and Layer aggregators. However, the authors presented a novel differentiable search algorithm. Cai et al. \cite{relatedcai} introduced a GNAS approach featuring a uniquely designed search space and a gradient-based search approach. The authors developed a three-level Graph Neural Architecture Paradigm (GAP) that includes two types of fine-grained atomic operations (neighbor aggregation and feature filtering) that are derived from message-passing, to build the search space. Li et al. \cite{relatedli} introduced an innovative dynamic one-shot search space designed for multi-branch neural architectures within GNNs. The dynamic nature of the search space offers a larger capacity than a larger predefined search space. The architectures with lower importance weights are removed periodically from the population, while the candidate operations are unique to every edge of the computational graph. The authors performed both supervised and unsupervised techniques for the training part.  Zhao et al. \cite{relatedzhao} proposed a gradient-based architecture search method for predicting a system's remaining useful life. Their approach models the search space as a directed acyclic graph (DAG), where nodes represent latent representations and edges represent transformation operations. By employing candidate operations like ReLU and tanh, along with the softmax function, they make the search space continuous and the objective function differentiable, facilitating gradient-based optimization methods to find the optimal architecture.

\subsection{Related work review findings}
From the aforementioned research works, it is clear that there have been many approaches to the task of bot detection. Previous studies include supervised, unsupervised, and graph neural network (GNN) based methods. While they have shown promising results, the relentless evolution of bot accounts toward simulating human-like patterns poses a significant challenge to their effectiveness. These models are constrained by fixed architectures, limiting their adaptability to newer datasets. 

Little work has been done in employing Neural Architecture Search methods in GNN-based methodologies for bot detection. Our work shifts the focus on overcoming the performance limitations due to fixed architectures, by utilizing DFG-NAS to search for the best configuration of Propagation and Transformation functions in the message passing protocol of our RGCNs. Instead of extensive feature engineering our model searches for the permutation with the highest accuracy and aims for better adaptability in newer datasets that will depict future bots' behavior. Moreover, DFG-NAS presents high advantages, as it is suitable for GNN-based methods and overcomes over-smoothing and model degradation issues with the gate and skip-connection operations.

\section{Dataset}
The TwiBot-20 Dataset \cite{twibotdataset} is a publicly available dataset, constructed with a breadth-first search (BFS) methodology. The dataset includes information about each user's profile information obtained from the Twitter API, recent tweets, and domains of the user's interest. It also contains information about the user's neighborhood, which helps us construct a heterogeneous graph from the following relationships.  Table 1 presents all the attributes of the TwiBot-20 Dataset and a short description of them. The information from the user profiles is further mentioned in the preprocessing part of the model. The graph that is constructed consists of 229,580 nodes and 227,979 edges. The objective of the bot detection system is to distinguish between bots and genuine users by analyzing information from user descriptions, tweets, numerical and categorical properties, as well as neighborhood information.

\begin{table}[t!]
        \centering
        \caption {TwiBot-20 Dataset Attributes}
        \begin{tabular}{ c | c } 
         \hline
         Attribute & Description \\ [0.5ex] 
         \hline
         ID & ID from Twitter to identify the user \\ 
         profile & profile information from Twitter API \\ 
         tweet & 200 recent tweets of the user \\ 
         neighbor & 20 random followers and followings of the user \\ 
         domain & domain of the user (politics, business, entertainment, sports) \\ 
         label & label of the user ('1': bot, '0':human)\\ [1ex] 
         \hline
        \end{tabular}        
\end{table}

\section{Methodology}
\begin{figure*}[!t]
\centering
\includegraphics[width=5.5in]{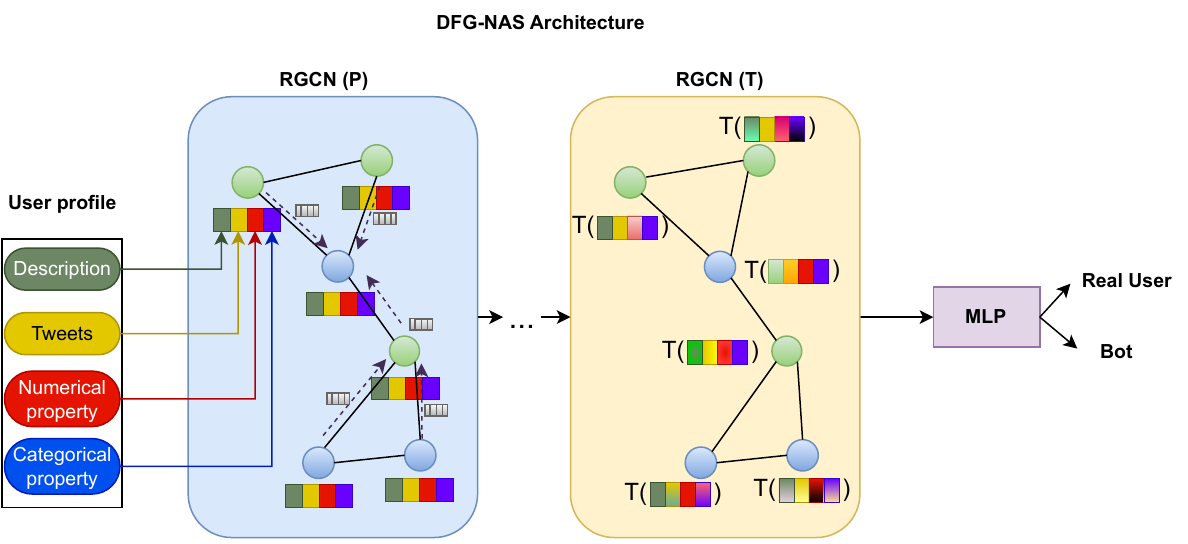}
\caption{Model used for Bot detection. User metadata is fed to the architecture proposed by NAS. The P step includes message aggregation from neighbour nodes. The T step includes the transformation process on each node based on neighbour relations. In the final part, an MLP decides whether the account belongs to a real user or a bot.}
\label{proposed_approach}
\end{figure*}

\begin{figure}
\centering
\includegraphics[width=3.5in]{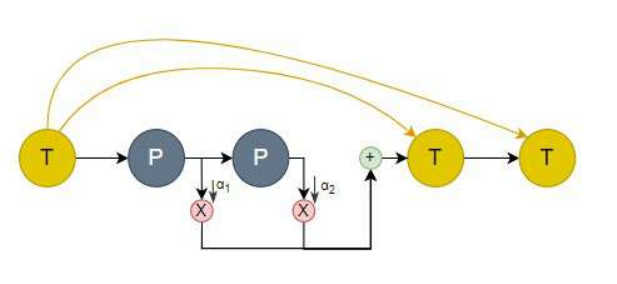}
\caption{Example of connections between the layers of NAS architecture. New T steps congregate information from all previous T steps. P steps propagate their embeddings and sum them up for the next T step.}
\label{example_nas}
\end{figure}

In this part, we present a complete analysis of our methodology. First, we describe the preprocessing of the user metadata used in our model. Next, we introduce the use of Relational Graph Convolutional Neural Networks and the two functions in Message Passing. Last, we explain the use of DFG-NAS \cite{dfgnas} in searching for the best permutation of Propagation and Transformation functions. In Figure~\ref{proposed_approach}, we depict the architecture of the model on a higher level, while Figure~\ref{example_nas} presents the connections between the different layers of an example configuration of P and T functions.

\subsection{Data Preprocessing}
We follow the preprocessing suggested by Feng et al. for BotRGCN \cite{botrgcn}. Each user's representation includes metadata that are preprocessed as follows:

\begin{itemize}
    \item \textbf{Overall}: User's description, tweets, numerical and categorical properties are encoded and concatenated to finally represent the user's metadata:
    \begin{equation}
        r = [r_b;r_t;r^{num}_p;r^{cat}_p] \in \mathbb{R}^{D \times 1}
        \label{overallfeatures}
    \end{equation}
    where $D$ is the user embedding dimension. Each feature's procession and representation are explained below. Later we will prove that the model's performance is attributed to all these features and not only to the heterogeneous graph.

    \item \textbf{User description}: The user descriptions are encoded with pre-trained RoBERTa:
    \begin{equation}
        \bar{b} = RoBERTa(\{b_i\}^L_{i=1}), \bar{b} \in \mathbb{R}^{D_s \times 1}
        \label{des1}
    \end{equation}
    where $\bar{b}$ denotes the user description representation and $D_s$ is the dimension of the RoBERTa embedding. The vectors for the user's description are derived:
    \begin{equation}
        r_b = \phi (W_B \cdot \bar{b} + b_B), r_b \in \mathbb{R}^{D/4\times 1}
        \label{des2}
    \end{equation}
     where $W_B$ and $b_B$ represent trainable parameters, $\phi$ denotes the activation function, and $D$ is the dimension of the embedding.

     \item \textbf{User tweets}: The user tweets are also encoded using RoBERTa. The ultimate representation of the user's tweets, denoted as $r_t$, is computed as the average of the representations of all individual tweets.

    \item \textbf{User numerical properties}: The user's numerical properties are adopted straight from the Twitter API with no feature engineering and presented in Table 2. For this information z-score normalization is conducted to get the representation $r^{num}_p$ from a fully connected layer.
        
     \begin{table}[t!]
        \centering
        \caption{User Numerical Properties}
        \begin{tabular}{ c | c } 
         \hline
         Feature Name & Description \\ [0.5ex] 
         \hline
         \#followers & number of followers \\ 
         \#followings & number of followings \\ 
         \#favorites & number of likes \\ 
         \#statuses & number of statuses \\ 
         active\_days & number of active days \\ 
         screen\_name\_length & screen name character count \\ [1ex] 
         \hline
        \end{tabular}        
    \end{table}

    \item \textbf{User categorical properties}: The user's categorical properties are also encoded with MLPs and GNNs, without feature engineering, just as the numerical properties. They are adopted straight from the Twitter API and presented in Table 3. After one-hot encoding, they are concatenated and transformed through a fully connected layer and leaky-relu to get their representation $r^{cat}_p$.

    \begin{table}[ht!]
    \scriptsize
        \centering
        \caption{User Categorical Properties}
        \begin{tabular}{ c | c } 
         \hline
         Feature Name & Description \\ [0.5ex] 
         \hline
         protected & protected or not \\ 
         geo\_enabled & geo-location enabled or not \\ 
         verified & verified or not \\ 
         contributors\_enabled & enable contributors or not \\ 
         is\_translator & is translator or not \\ 
         is\_translation\_enabled & translation or not \\ 
         profile\_background\_tile & the background tile \\ 
         profile\_user\_background\_image & background image or not \\ 
         has\_extended\_profile & extended profile or not \\ 
         default\_profile & the default profile \\ 
         default\_profile\_image & the default profile image \\ [1ex] 
         \hline
        \end{tabular}
    \end{table}
\end{itemize}

\subsection{Relational Graph Convolutional Neural Networks}
Our method builds a heterogeneous graph out of the following relationships. Users are considered nodes and the 'following' and 'followers' relations are represented as edges connecting the nodes. The user's 'followers' are therefore represented differently than the user's 'following'. The heterogeneous graph that is constructed can represent better the relations between users and more relations between the users could be integrated into the graph if supported by the dataset. The users also contain the concatenated metadata that we described below.

To combine the users' representations with the relationships between users we make use of RGCNs. The message-passing process in RGCNs comprises two fundamental operations: propagation (P) of the representations of the user's neighbors and transformation (T) on these representations. Below we describe the process behind the two functions:

\begin{itemize}
\item \textbf{Propagation (P)}: Propagation includes message aggregation from neighbour nodes without explicit node feature transformation. The mathematical expression for the propagation step is as follows:

\begin{equation}
    h_{i}^{(l+1)} = \sum_{r \in R} \sum_{j \in N_{i}^{r}} \frac{1}{c_{i,r}} W_{r}^{(l)} h_{j}^{(l)}
    \label{propagationrgcn}
\end{equation}

where $h_{i}^{(l+1)}$ is the new node feature after propagation, R is the set of relations, $N_{i}^{r}$ are the neighbors of the node with relation r, $c_{i,r}$ is a normalization constant that can be learned or chosen in advance (for example $c_{i,r}$ = $N_{i}^{r}$) and $W_{r}^{(l)}$ is the learnable weight matrix for relation r.

\item \textbf{Transformation (T)}: Transformation occurs on each node based on the relations. The mathematical expression for the transformation step is as follows:

\begin{equation}
    h_{i}^{(l+1)} = W_{root} h_{i}^{(l)} + \sum_{r \in R}(W_{r} h_{i}^{(l)})
    \label{transformationrgcn}
\end{equation}

where $h_{i}^{(l+1)}$ is the new node feature after transformation, $W_{root}$ is the learnable weight matrix for the root node, $W_{r}$ is the learnable weight matrix for relation r and R is the set of relations.
\end{itemize}

We segregate these two types of functions since combinations of them will construct the search space for the architecture search.

\subsection{Graph Neural Architecture Search}
The use of Graph Neural Networks offers undeniable advantages in the task of bot detection. However, maximizing their performance may require extensive feature engineering. This is why we employ Graph Neural Architecture Search, using the model DFG-NAS \cite{dfgnas}. Thus, we search for the permutation of Propagation and Transformation steps that achieves the highest accuracy. Most G-NAS methods have a fixed pipeline length since the performance decreases with too many P operations as the layers become deeper, which is referred to as the over-smoothing issue. Propagation and transformation operations regulate the effect of smoothing. Moreover, with unlimited pipeline length DFG-NAS searches for more flexible pipelines of P and T operations, using an evolutionary algorithm. It also makes use of gating and skip-connection mechanisms in the P and T operations, respectively.

The search space includes P-T combinations and the number of P-T operations. The output of node $v$ in the l-th layer is represented by $o^{(l)}_v$ in a single P or T operation within a single GNN layer of the model. The layer indices of all P and T operations are included in two sets, $L_P$ and $L_T$. The connections of P and T are depicted in Figure 1(b) and also described below:

\textbf{Propagation connections}: An imminent problem in GNNs is over-smoothing or under-smoothing, a problem that arises with too many or too few propagation operations. To achieve suitable smoothness for different nodes, the P operations are amplified with a gating mechanism. If the next operation is also P, the result of the l-th P operation is the propagated node embedding of $o^{(l-1)}$. On the other hand, if T is the next operation, a node-adaptive combination weight is allocated for the node embeddings propagated by all of the previous P operations. Formulatively:

\begin{equation}
    z^{(l)}_v = P(o^{(l-1)}_v)
    \label{prop1}
\end{equation}
\begin{equation}
    o^{(l)}_v = \begin{cases} 
      z^{(l)}_v, & followed\ by\ P \\
      \displaystyle\sum\limits_{i \in L_P, i \leq l} softmax(a_i) z^{(i)}_v, & followed\ by\ T \\
   \end{cases}
    \label{prop2}
\end{equation}

where $a_i = \sigma(s \cdot o^i_v)$ represents the weight for the i-th layer output of node $v$. Here, $s$ is the learnable vector shared among the entirety of nodes, and $\sigma$ denotes the Sigmoid function. To ensure proper scaling, the Softmax function is employed to normalize the sum of gating scores, making it equal to 1.

\textbf{Transformation connections}: An imminent issue with GNNs is the model degradation issue, caused by a hyperbolic amount of transformation operations and may result in a reduction of the model's accuracy. To mitigate this issue, skip-connection mechanisms are used in T operations. Each T operation's input is the total of all the T operations' outputs up to the last layer and the output from the layer before it. The input and output of the l-th T operation can be formulated as:

\begin{equation}
    z^{(l)}_v = o^{(l-1)}_v + \displaystyle\sum\limits_{i \in L_T, i < m(l)} o^{(i)}_v
    \label{tran1}
\end{equation}
\begin{equation}
    o^{(l)}_v = \sigma(z^{(l)}_v w^{(l)})
    \label{tran2}
\end{equation}

where $m(l)$ represents the index of the last T operation before the l-th layer, and $W(l)$ denotes the trainable parameter in the l-th T operation.

Evolutionary algorithms are a class of optimization algorithms inspired by biological evolution that aim to achieve the best accuracy in offspring through mutations. In our case, each GNN architecture is represented as a sequence of P and T operations. Each pipeline can be considered a chromosome and the mutations that occur simulate nature's mutations. These mutations can happen at any random position in the sequence. In our instance, four different cases of mutation can be enforced:

\begin{itemize}
    \item \textbf{+P}: append a propagation operation
    \item \textbf{+T}: append a transformation operation
    \item \textbf{P→T}: replace a propagation operation with a transformation one
    \item \textbf{T→P}: replace a transformation operation with a propagation one
\end{itemize}

Initially, k distinct GNN designs are generated at random and evaluated on the validation set. These architectures represent the initial population set Q. Subsequently, m (m $<$ k) members of the population are randomly sampled, and parent A is determined by selecting the member with the highest validation accuracy. By enforcing a random mutation of the four presented on A, a child architecture B is produced. B is then evaluated and added to the population, and the oldest person is eliminated. After T generations of this procedure, the architecture with the best performance is eventually returned.

DFG-NAS returns a sequence of P and T steps. As illustrated in Figure 1(a), each step consists of an RGCN that conducts one of the two main functions as we described incorporating both the user metadata and the user relations. After the RGCNs layers an MLP is employed to finally distinguish bots from genuine users.

\section{Experiments}
\subsection{Baselines}
 We compare our proposed apporach to the state-of-the-art models that are referenced in the paper of BotRGCN \cite{botrgcn}. These experiments are all ran on the same dataset as the one we used for a fair comparison. We are using the published results for the comparison. More specifically, we compare our model to these state-of-the-art models:
\begin{itemize}
    \item Lee et al. \cite{relatedlee} employed different supervised algorithms with several user features.
    \item Yang et al. \cite{relatedyang} used a combination of supervised and unsupervised learning with minimal user features.
     \item Kudugunta et al. \cite{relatedkuduganta} used both the tweets and the account metadata.
     \item Wei et al. \cite{relatedwei} employed an RNN model utilizing only the user's tweets.
     \item Miller et al. \cite{relatedmiller} extracted 107 features and employed stream clustering algorithms.
     \item Cresci et al. \cite{relatedcresci} identified bots by computing the longest common substring between encoded sequences of users.
     \item Botometer \cite{relatedbotometer} is a web-based program that leverages more than 1,000 user features.
     \item Alhosseini et al. \cite{relatedalhosseini} introduced graph convolutional neural networks in bot detection.
     \item SATAR \cite{relatedsatar} leverages the user's semantics, property, and neighborhood information
     \item BotRGCN et al. \cite{botrgcn} used the user's description, tweets, numerical and categorical properties, and neighborhood information.
     \item Ilias et al. \cite{relatedilias} designed two cross-attention layers based on the digital DNA sequence. 
\end{itemize}

\subsection{Experiment Settings}
The experiment was run on Google Colab using Nvidia's T4 GPUs. The population set k for the architectural search is 15, and the maximum generation time T is 80. The training budget of each GNN architecture is 70 epochs. These numbers although limited due to our resources, provide a great example of the efficiency of our model. More complex architectures that we tested do not necessarily provide better results. Also, the number of epochs is sufficient to get a good idea of each architecture's accuracy. Adam optimizer is used for training, and its learning rate is set to 0.04. The criterion is Cross Entropy Loss and the regularization factor is 2e-4. Dropout is applied to all feature vectors at a rate of 0.5, and dropout among GNN layers is set to 0.8.

After running the NAS method we process the results and examine the five architectures with the best accuracy in the validation set. Each architecture is now trained with 100 epochs on the TwiBot-20 dataset \cite{twibotdataset}. The train set is 70\% of the dataset, the validation set is 20\% and the test set is 10\%. Adam optimizer with a learning rate of 1e-3 is also used for training. Then each architecture is tested on the test set. We will present the findings of these experiments below.

\subsection{Evaluation Metrics}
We assess our model's performance using its Accuracy, F1-score, Precision, Recall, Specificity, and MCC. These metrics are computed by labeling the bots as the positive class and the genuine users as the negative class. To compare the performance of our model to the other baseline models we will only use the metrics Accuracy, F1-score, and MCC.

\section{Results}
Each architecture during the search is saved with its P-T configuration, accuracy in the validation set, and accuracy in the test set. In Figure 2, the five architectures with the highest validation accuracy that are chosen from the NAS method are depicted.

\begin{figure}[!t]
\centering
\includegraphics[width=3.4in]{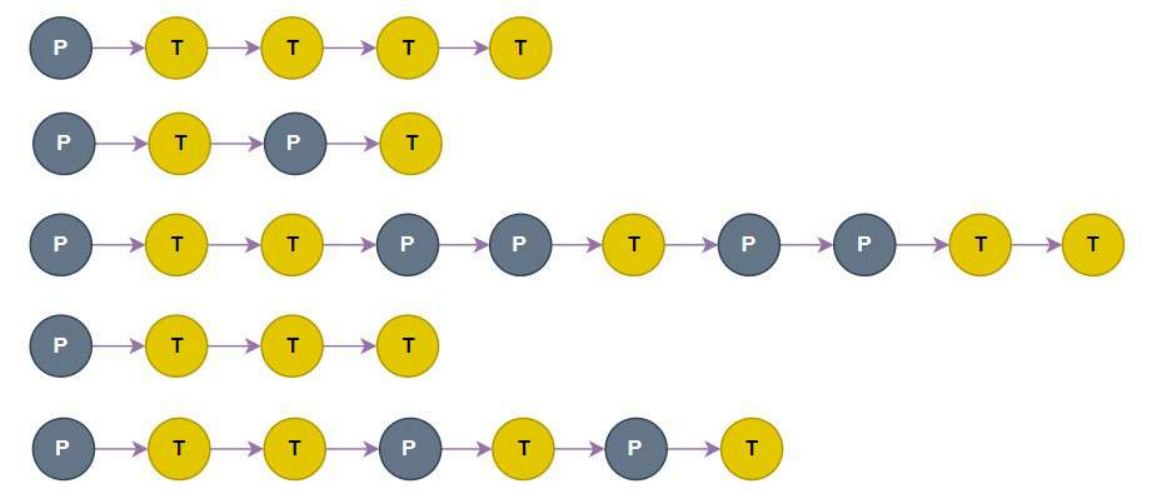}
\caption{Permutations of Propagation (P) and Transformation (T) functions of the top-5 performing architectures from DFG-NAS. Their validation accuracies in the architecture search (from up to down) are: 87.01\%, 86.99\%, 86.95\%, 86.89\%, 86.82\%}
\label{fivearchs}
\end{figure}

These architectures are trained and tested from scratch in TwiBot-20 dataset. We present all the metrics attained by all the architectures in Table 4.

\begin{table*}[ht!]
\centering
\caption{\textbf{Performance of the architectures from architecture search.} Values are reported as mean ± standard deviation. Five runs of results are averaged. The best outcomes for each evaluation metric are in bold.}

\begin{tabular}{| c || c | c | c | c | c | c | c |} 
 \hline
 Model & Accuracy & F1-score & Precision & Recall & Specificity & MCC \\ [0.5ex] 
 \hline
 1st Architecture & 0.852 $\pm$ 0.005 & 0.865 $\pm$ 0.008 & 0.851 $\pm$ 0.015 & 0.880 $\pm$ 0.031 & 0.818 $\pm$ 0.027 & 0.702 $\pm$ 0.010 \\ 
 
 2nd Architecture & 0.855 $\pm$ 0.004 & 0.869 $\pm$ 0.005 & 0.853 $\pm$ 0.007 & 0.886 $\pm$ 0.012 & 0.819 $\pm$ 0.012 & 0.709 $\pm$ 0.009 \\ 
 
 3rd Architecture & \textbf{0.857 $\pm$ 0.004} & \textbf{0.871 $\pm$ 0.003} & 0.849 $\pm$ 0.008 & \textbf{0.895 $\pm$ 0.007} & 0.812 $\pm$ 0.013 & \textbf{0.712 $\pm$ 0.007} \\ 
 
 4th Architecture & 0.852 $\pm$ 0.006 & 0.864 $\pm$ 0.008 & 0.856 $\pm$ 0.009 & 0.873 $\pm$ 0.026 & 0.828 $\pm$ 0.018 & 0.702 $\pm$ 0.013 \\ 
 
 5th Architecture & 0.852 $\pm$ 0.007 & 0.864 $\pm$ 0.008 & \textbf{0.858 $\pm$ 0.003} & 0.872 $\pm$ 0.019 & \textbf{0.829 $\pm$ 0.007} & 0.703 $\pm$ 0.014 \\ 
 [0.5ex] 
 \hline
\end{tabular}
\end{table*}

All selections achieve good metrics and present advantages in bot detection over state-of-the-art methods. These results underscore the significant advantages that emerge from employing architecture search techniques regarding the field of bot recognition. Moreover, they establish the efficiency of utilizing user features and relationships between users in bot detection.

Upon closer examination of the results, the third architecture achieves the best evaluation metrics. The fifth architecture has the highest precision. However, all the architectures present high metrics of accuracy, F1-score, and MCC and whichever architecture we choose could compete with state-of-the-art models. From now on we will refer to the third architecture as our model, since it provides the highest accuracy.

In Table 5 we present the performance of the baseline methods on the TwiBot-20 dataset compared to ours. We see that our model benefits from the search for the fittest architecture that we performed beforehand, as it achieves a higher accuracy, F1-score, and MCC than other state-of-the-art methods.

\begin{table*}[ht!]
\centering
\caption{\textbf{Performance of models on the TwiBot-20 dataset.} Values are reported as mean ± standard deviation. Five runs of results are averaged. The best outcomes for each evaluation metric are in bold.}
\begin{tabular}{| c || c | c | c |} 
 \hline
 Model & Accuracy & F1-score & MCC \\ [0.5ex] 
 \hline
 \cite{relatedlee} &  0.7456 & 0.7823  & 0.4879 \\
 \cite{relatedyang} & 0.8191 & 0.8546 & 0.6643 \\
 \cite{relatedkuduganta} & 0.8174 & 0.7517 & 0.6710 \\ 
 \cite{relatedwei} & 0.7126 & 0.7533 & 0.4193 \\ 
  \cite{relatedmiller} & 0.4801 & 0.6266 & -0.1372 \\
 \cite{relatedcresci} & 0.4793 & 0.1072 & 0.0839 \\
 \cite{relatedbotometer} & 0.5584 & 0.4892 & 0.1558 \\
 \cite{relatedalhosseini} & 0.6813 & 0.7318 & 0.3543 \\
 \cite{relatedsatar} & 0.8412 & 0.8642 & 0.6863 \\ 
 \cite{botrgcn} & 0.8462 & 0.8707 & 0.7021 \\
 \cite{relatedilias} & 0.7466 & 0.7630 & -- \\
 
 \textbf{ours} & \textbf{0.8568 $\pm$ 0.004} & \textbf{0.8712 $\pm$ 0.003} & \textbf{0.7116 $\pm$ 0.007} \\
 [0.5ex] 
 \hline
\end{tabular}
\end{table*}

\section{Ablation Study}
To demonstrate our model's effectiveness and integrity we will perform an ablation study on the basic ideas: the user's features used for the training, the Gate operation, and the skip-connection operation.

To prove that using multi-modal information is vital to our model performance we will train the architecture that produces the best results with reduced features. We will reduce one feature at a time and use combinations of the features for the training. We present the results in Table 6.

\begin{table*}[ht!]
\scriptsize
\centering
\caption{\textbf{Training model with less features}. Values are reported as mean ± standard deviation. Five runs of results are averaged. The best outcomes for each evaluation metric are in bold.}
\begin{tabular}{| c || c | c | c | c | c | c | c |}
 \hline
 Model & Accuracy & F1-score & Precision & Recall & Specificity & MCC \\ [0.5ex] 
 \hline
 Ours & 0.857 $\pm$ 0.004 & 0.871 $\pm$ 0.003 & 0.849 $\pm$ 0.008 & 0.895 $\pm$ 0.007 & 0.812 $\pm$ 0.013 & 0.712 $\pm$ 0.007 \\ 
 
 w/o description & \textbf{0.859 $\pm$ 0.004} & \textbf{0.875 $\pm$ 0.004} & 0.845 $\pm$ 0.002 & 0.906 $\pm$ 0.008 & 0.804 $\pm$ 0.004 &\textbf{0.718 $\pm$ 0.008} \\ 
 
 w/o tweets & 0.833 $\pm$ 0.007 & 0.858 $\pm$ 0.007 & 0.796 $\pm$ 0.004 & 0.93 $\pm$ 0.013 & 0.719 $\pm$ 0.005 &0.671 $\pm$ 0.016 \\ 
 
 w/o numerical & \textbf{0.859 $\pm$ 0.003} & 0.872 $\pm$ 0.005 & \textbf{0.856 $\pm$ 0.012} & 0.889 $\pm$ 0.023 & \textbf{0.823 $\pm$ 0.021} &0.716 $\pm$ 0.007 \\ 
 
 w/o categorical & 0.792 $\pm$ 0.003 & 0.814 $\pm$ 0.001 & 0.791 $\pm$ 0.010 & 0.840 $\pm$ 0.014 & 0.738 $\pm$ 0.021 &0.582 $\pm$ 0.005 \\ 

 des + tweets & 0.759 $\pm$ 0.007 & 0.773 $\pm$ 0.009 & 0.789  $\pm$ 0.022 & 0.758 $\pm$ 0.034 & 0.761 $\pm$ 0.045 & 0.519 $\pm$ 0.014 \\ 

 cat + num & 0.817 $\pm$ 0.001 & 0.855 $\pm$ 0.001 & 0.749 $\pm$ 0.001 & \textbf{0.996 $\pm$ 0.001} & 0.607 $\pm$ 0.002 & 0.668 $\pm$ 0.002 \\ 
 [0.5ex] 
 \hline
\end{tabular}
\end{table*}

We see that training with reduced features may achieve higher metrics in some cases. Notably, training without descriptions has a higher F1-score than the original model but has a lower precision. Also, training without tweets has a higher recall value. Training without numerical properties has a higher precision and specificity but a lower MCC than training without description. Training with only the categorical and numerical properties has the highest recall. Therefore, training with combinations of features does not achieve as high metrics as training with all the features in each case, meaning that all features contribute to the model's performance. These remarks are important to consider for future research in ensuring the dataset's quality, but training the model with all the features provided makes it more adaptable to other datasets. For further understanding we will train the model using only one feature at a time, to investigate their importance separately. We present the results in Table 7.

\begin{table*}[ht!]

\centering
\caption{\textbf{Training model with only one feature}. Values are reported as mean ± standard deviation. Five runs of results are averaged. The best outcomes for each evaluation metric are in bold.}
\begin{tabular}{| c || c | c | c | c | c | c | c |} 
 \hline
 Model & Accuracy & F1-score & Precision & Recall & Specificity & MCC \\ [0.5ex] 
 \hline
 Ours & \textbf{0.857 $\pm$ 0.004} & \textbf{0.871 $\pm$ 0.003} & \textbf{0.849 $\pm$ 0.008} & 0.895 $\pm$ 0.007 & \textbf{0.812 $\pm$ 0.013} & \textbf{0.712 $\pm$ 0.007} \\
 
 only description & 0.699 $\pm$ 0.007 & 0.74 $\pm$ 0.008 & 0.695 $\pm$ 0.015 & 0.793 $\pm$ 0.033 & 0.589 $\pm$ 0.046 & 0.392 $\pm$ 0.014 \\
 
 only tweets & 0.585 $\pm$ 0.011 & 0.643 $\pm$ 0.017 & 0.602 $\pm$ 0.008 & 0.691 $\pm$ 0.037 & 0.461 $\pm$ 0.033 & 0.157 $\pm$ 0.022 \\
 
 only numerical & 0.679 $\pm$ 0.02 & 0.758 $\pm$ 0.013 & 0.641 $\pm$ 0.018 & 0.929 $\pm$ 0.034 & 0.385 $\pm$ 0.063 & 0.383 $\pm$ 0.039 \\
 
 only categorical & 0.817 $\pm$ 0.001 & 0.853 $\pm$ 0.001 & 0.747 $\pm$ 0.001 & \textbf{1.000 $\pm$ 0.001} & 0.6 $\pm$ 0.001 &0.667 $\pm$ 0.001 \\
 [0.5ex] 
 \hline
\end{tabular}
\end{table*}

Obviously, the model trained with all the features has the best performance. From the results, we deduce that the categorical property is the feature that contributes the most to the model's sufficient accuracy. This ablation study proves that all features are advantageous for training our model to perform well in the task of bot detection. However, they do not contribute equally, and more studies to enhance the quality of the datasets could benefit future studies of bot detection.

Next, we compare the architecture that results from the architecture search with a Gate operation and without a Gate operation. The findings of this ablation study are depicted in Table 8. We see that the architecture without the gate has a reduced accuracy by 0.5\% compared to the model's and a reduced F1-score by 0.46\%. The gating mechanism dynamically consolidates information from all propagation steps, effectively regulating the smoothness of various nodes. Without it, the T operations take as input only the last output of the P steps. This is the reason the model underperforms without the Gate operation in the P functions, as it may suffer from over-smoothing. The architectures that are examined during this search have more T steps and shallower propagation processes, failing to obtain information from nodes during message passing as successfully as the original model. This ablation study proves the importance of the Gate operation in the P functions during our architecture search. 

\begin{table*}[ht!]
\centering
\caption{\textbf{Ablation study on Gate operation}. Values are reported as mean ± standard deviation. Five runs of results are averaged. The best outcomes for each evaluation metric are in bold.}
\begin{tabular}{| c || c | c | c | c | c | c | c |} 
 \hline
 Model & Accuracy & F1-score & Precision & Recall & Specificity & MCC \\ [0.5ex] 
 \hline
 With Gate & \textbf{0.857 $\pm$ 0.004} & \textbf{0.871 $\pm$ 0.003} & \textbf{0.849 $\pm$ 0.008} & \textbf{0.895 $\pm$ 0.007} & \textbf{0.812 $\pm$ 0.013} & \textbf{0.712 $\pm$ 0.007} \\
 
 Without Gate & 0.853 $\pm$ 0.003 & 0.867 $\pm$ 0.004 & 0.845 $\pm$ 0.010 & 0.891 $\pm$ 0.016 & 0.808  $\pm$ 0.018 & 0.704 $\pm$ 0.007 \\
 [0.5ex] 
 \hline
\end{tabular}
\end{table*}

Finally, we compare the architecture that results from the architecture search with a skip-connection operation and without a skip-connection operation. The findings of this ablation study are depicted in Table 9. We see that the architecture without the gate has a reduced accuracy by 0.93\% compared to the model's and a reduced F1-score by 1.2\%. Without the skip-connection operation, the input of the T steps is only the output of the last step. This may lead to the degradation of the model as the transformation functions can increase. The processing of the messages from nodes is not as effective and the accuracy declines. This ablation study proves the importance of the skip-connection operation in the T functions during our architecture search.

\begin{table*}[ht!]
\centering
\caption{\textbf{Ablation study on skip-connection operation}. Values are reported as mean ± standard deviation. Five runs of results are averaged. The best outcomes for each evaluation metric are in bold.}

\begin{tabular}{| c || c | c | c | c | c | c | c |} 
 \hline
 Model & Accuracy & F1-score & Precision & Recall & Specificity & MCC \\ [0.5ex] 
 \hline
 With skip & \textbf{0.857 $\pm$ 0.004} & \textbf{0.871 $\pm$ 0.003} & 0.849 $\pm$ 0.008 & \textbf{0.895 $\pm$ 0.007} & 0.812 $\pm$ 0.013 & \textbf{0.712 $\pm$ 0.007} \\
 
 Without skip & 0.849 $\pm$ 0.009 & 0.860 $\pm$ 0.01 & \textbf{0.857 $\pm$ 0.010} & 0.863 $\pm$ 0.026 & \textbf{0.831 $\pm$ 0.017} & 0.695 $\pm$ 0.018 \\
 [0.5ex] 
 \hline
\end{tabular}
\end{table*}

\section{Discussion}

\subsection{Implications}
 The proliferation of social media bots has prompted concerns regarding user safety and their broader societal impact. Bot detection, a focal point of contemporary studies, is not only explored through the lens of machine learning but also delves into the realms of social science. Various methodologies have been employed, encompassing supervised or unsupervised learning or a hybrid of both. A relatively recent and innovative approach involves Graph Neural Network (GNN)-based architectures, integrating diverse user features and interactions to construct a comprehensive graph representation. In our work, we formulate a heterogeneous graph that captures the following relationships between users, incorporating nodes with information on user profiles, tweets, and interests. This novel contribution enhances existing bot detection research by demonstrating the efficacy of integrating and analyzing user relationships.

As technology advances, the adaptive nature of bots poses an ongoing challenge for detection models, rendering many state-of-the-art architectures ineffective against newer datasets. The pressing need for adaptable models underscores the importance of overcoming the limitations associated with fixed architectures. Neural Architecture Search (NAS) models prove to be a promising solution, demonstrating their potential to enhance model efficiency in real-world tasks by automatically searching through various architectures. Historically, the adoption of NAS techniques for bot detection is limited, so we propose the implementation of an adapted DFG-NAS. By integrating DFG-NAS and tailoring it to Relational Graph Convolutional Networks, we explore optimal permutations of Propagation and Transformation steps in the message-passing protocol of the RGCN layers. Our investigation showcases superior performances of the top architectures compared to state-of-the-art models. Our work is one of the starting points in implementing architecture search models on bot detection. Our research findings encourage further exploration into how NAS models can automatically construct more effective architectures, resulting in a future restraint of the existence of bots.

\subsection{Applicability of our Approach to Different Types of Social Interaction}

In this section, we examine the applicability of our introduced approach to other types of social interaction besides social media.

\begin{itemize}
    \item \textbf{Online Gaming:} Bots impersonate human players to manipulate game outcomes. Bots are capable of playing without breaks. Therefore, they are able to gather resources, items, and so on very quickly which help them go to the next stage of gaming \cite{KANG20131384}. Thus, people end playing with bots; so, it is impossible to win them. This fact entails serious issues, i.e., unfair gaming. Therefore, the early detection of bots in gaming is crucial, in order to ensure fair play in competitive and multiplayer games. Our method could be adapted by using response times, movement patterns, and time-series data as input features. 
    \item \textbf{Customer Reviews and Rating Platforms:} Bots are often used for creating fake reviews and inflating rating in review platforms, including Amazon and Yelp. The main aim of bots is to promote specific products, restaurants, and so on. Our approach could be easily adapted to this case, since textual, timing, and user behaviour features will be used.
    \item \textbf{Digital Voting and Polling Systems:} Bots are used to alter the results of Internet Polling \cite{5555837}. Therefore, early recognition of bots in voting is crucial, so as to ensure reliable outcomes. Our method can be adapted by integrating features, such as IP addresses, voting patterns, and timing. 
    \item \textbf{Email and Messaging Systems:} Bots are responsible for spam and phishing. Early detection of bots is crucial for enhancing security. Features, including email headers, IP addresses, etc., must be incorporated in our study.  
\end{itemize}

\subsection{Limitations}

Our study comes with some limitations. Firstly, we conducted our experiments only on one dataset, which does not ensure generalizabilty of our proposed approach. Therefore, in the future, we aim to test our method on TwiBot-22 dataset \cite{10.5555/3600270.3602825}. Secondly, our method is based on the collection of labelled data. Obtaining labelled data is a difficult task. For this reason, unsupervised and self-supervised learning algorithms have been developed for addressing the issue of labels' scarcity. Applying unsupervised and self-supervised learning in conjunction with our approach is one of our future plans. Thirdly, we did not tune the hyperparameters due to limited access to GPU resources. Hyperparameter tuning ensures that optimal performance is obtained. 
Finally, we represented each user as a concatenation of features. Concatenation does not capture the inherent correlation of the different modalities. In the future, we aim to use multimodal fusion methods for constructing each user's representation \cite{9926818, ILIAS2023110834, dartsad}.

\section{Conclusions and Future Work}
As social media continues to play a pivotal role in shaping public opinion and discourse, the development of effective and adaptive bot detection methods becomes increasingly crucial for maintaining the integrity and trustworthiness of online information.
In this study, we introduced a novel model for identifying bots, integrating GNNs and NAS algorithms, demonstrating significant performance gains. The integration of Graph Neural Architecture Search empowered us to dynamically determine optimal combinations of propagation and transformation operations in the graph neural network architecture. This adaptive architecture effectively addresses the constraints imposed by fixed structures, introducing a level of flexibility essential for improving the performance on the bot detection task.  From the experiment results we conclude that the five architectures with the highest validation accuracy, during the architecture search, are quite efficient in our task and compete with other models. Meanwhile, the one with the highest accuracy achieves a test accuracy of 85,68\%, surpassing other state-of-the-art models for bot detection. The outcomes of the experiment present promising prospects for integrating more Neural Architecture Search (NAS) methods into the domain of bot detection in various social media platforms.

The exploration of dynamic graph adaptations stands as a crucial avenue for future research in the task of bot identification in social media platform X. The dynamic nature of social networks, characterized by the continuous incorporation of new users, necessitates the development of mechanisms to seamlessly integrate these additions into the evolving graph structure. Investigating methods for real-time graph updates and exploring how the model adapts to the inclusion of new users will enhance the system's agility in capturing emerging bot behaviors within the dynamic social landscape. Furthermore, the prospect of transferring our model to other social media platforms emerges as a key future avenue. Extending the applicability of our approach beyond X involves understanding the unique dynamics and characteristics of different platforms. Future work should focus on developing a transferable framework capable of recognizing bot-like behaviors across diverse social networks. By addressing the nuances and variations in user interactions and content features, we can contribute to the development of a versatile bot detection system with broader applications in the ever-expanding realm of social media platforms.

\bibliographystyle{unsrt}  
\bibliography{references} 

\begin{thebibliography}{10}

\bibitem{ILIAS2023100270}
Loukas Ilias and Dimitris Askounis.
\newblock Multitask learning for recognizing stress and depression in social media.
\newblock {\em Online Social Networks and Media}, 37-38:100270, 2023.

\bibitem{10154134}
Loukas Ilias, Spiros Mouzakitis, and Dimitris Askounis.
\newblock Calibration of transformer-based models for identifying stress and depression in social media.
\newblock {\em IEEE Transactions on Computational Social Systems}, 11(2):1979--1990, 2024.

\bibitem{kerasiotis2024}
Marios Kerasiotis, Loukas Ilias, and Dimitris Askounis.
\newblock Depression detection in social media posts using transformer-based models and auxiliary features.
\newblock {\em Social Network Analysis and Mining}, 14:196, 2024.

\bibitem{introrussian}
Izzat Alsmadi and Michael~J. O'Brien.
\newblock How many bots in russian troll tweets?
\newblock {\em Information Processing \& Management}, 57(6):102303, 2020.

\bibitem{introtrolling}
Joshua Uyheng, J.D. Moffitt, and Kathleen~M. Carley.
\newblock The language and targets of online trolling: A psycholinguistic approach for social cybersecurity.
\newblock {\em Information Processing \& Management}, 59(5):103012, 2022.

\bibitem{introstrategy}
Yaozeng Zhang, Jing Ma, and Fanshu Fang.
\newblock How social bots can influence public opinion more effectively: Right connection strategy.
\newblock {\em Physica A: Statistical Mechanics and its Applications}, 633:129386, 2024.

\bibitem{introgatekeeping}
Yi~Xu, Deru Zhou, and Wei Wang.
\newblock Being my own gatekeeper, how i tell the fake and the real – fake news perception between typologies and sources.
\newblock {\em Information Processing \& Management}, 60(2):103228, 2023.

\bibitem{introknowledge}
Yantian Mi and Oberiri~Destiny Apuke.
\newblock How does social media knowledge help in combating fake news? testing a structural equation model.
\newblock {\em Thinking Skills and Creativity}, page 101492, 2024.

\bibitem{relatedlee}
Kyumin Lee, Brian Eoff, and James Caverlee.
\newblock Seven months with the devils: A long-term study of content polluters on twitter.
\newblock {\em Proceedings of the International AAAI Conference on Web and Social Media}, 5(1):185--192, Aug. 2021.

\bibitem{relatedcresci}
Stefano Cresci, Roberto Di~Pietro, Marinella Petrocchi, Angelo Spognardi, and Maurizio Tesconi.
\newblock Social fingerprinting: detection of spambot groups through dna-inspired behavioral modeling.
\newblock {\em IEEE Transactions on Dependable and Secure Computing}, page 1–1, 2017.

\bibitem{relatedalauthman}
Mohammad Alauthman, Nauman Aslam, Mouhammd Al-kasassbeh, Suleman Khan, Ahmad Al-Qerem, and Kim-Kwang {Raymond Choo}.
\newblock An efficient reinforcement learning-based botnet detection approach.
\newblock {\em Journal of Network and Computer Applications}, 150:102479, 2020.

\bibitem{botrgcn}
Shangbin Feng, Herun Wan, Ningnan Wang, and Minnan Luo.
\newblock Botrgcn: Twitter bot detection with relational graph convolutional networks.
\newblock In {\em Proceedings of the 2021 IEEE/ACM International Conference on Advances in Social Networks Analysis and Mining}, ASONAM ’21, page 236–239. ACM, 11 2021.

\bibitem{dfgnas}
Wentao Zhang, Zheyu Lin, Yu~Shen, Yang Li, Zhi Yang, and Bin Cui.
\newblock Deep and flexible graph neural architecture search.
\newblock In Kamalika Chaudhuri, Stefanie Jegelka, Le~Song, Csaba Szepesvari, Gang Niu, and Sivan Sabato, editors, {\em Proceedings of the 39th International Conference on Machine Learning}, volume 162 of {\em Proceedings of Machine Learning Research}, pages 26362--26374. PMLR, 17--23 Jul 2022.

\bibitem{twibotdataset}
Shangbin Feng, Herun Wan, Ningnan Wang, Jundong Li, and Minnan Luo.
\newblock Twibot-20: A comprehensive twitter bot detection benchmark.
\newblock In {\em Proceedings of the 30th ACM International Conference on Information \& Knowledge Management}, CIKM '21, page 4485–4494, New York, NY, USA, 2021. Association for Computing Machinery.

\bibitem{intro1}
Alessandro Bessi and Emilio Ferrara.
\newblock Social bots distort the 2016 u.s. presidential election online discussion.
\newblock {\em First Monday}, 21, 11 2016.

\bibitem{intro2}
Emilio Ferrara.
\newblock What types of covid-19 conspiracies are populated by twitter bots?
\newblock {\em First Monday}, 5 2020.

\bibitem{introecho}
Christian Scheibenzuber, Laurentiu-Marian Neagu, Stefan Ruseti, Benedikt Artmann, Carolin Bartsch, Montgomery Kubik, Mihai Dascalu, Stefan Trausan-Matu, and Nicolae Nistor.
\newblock Dialog in the echo chamber: Fake news framing predicts emotion, argumentation and dialogic social knowledge building in subsequent online discussions.
\newblock {\em Computers in Human Behavior}, 140:107587, 2023.

\bibitem{introcyber}
Tanjim Mahmud, Michal Ptaszynski, Juuso Eronen, and Fumito Masui.
\newblock Cyberbullying detection for low-resource languages and dialects: Review of the state of the art.
\newblock {\em Information Processing \& Management}, 60(5):103454, 2023.

\bibitem{rosgas}
Yingguang Yang, Renyu Yang, Yangyang Li, Kai Cui, Zhiqin Yang, Yue Wang, Jie Xu, and Haiyong Xie.
\newblock Rosgas: Adaptive social bot detection with reinforced self-supervised gnn architecture search.
\newblock {\em ACM Transactions on the Web}, 17(3):1–31, May 2023.

\bibitem{relatedkuduganta}
Sneha Kudugunta and Emilio Ferrara.
\newblock Deep neural networks for bot detection.
\newblock {\em Information Sciences}, 467:312–322, October 2018.

\bibitem{relatedwei}
Feng Wei and Uyen~Trang Nguyen.
\newblock Twitter bot detection using bidirectional long short-term memory neural networks and word embeddings.
\newblock In {\em 2019 First IEEE International Conference on Trust, Privacy and Security in Intelligent Systems and Applications (TPS-ISA)}, pages 101--109, 2019.

\bibitem{relatedcai}
Shaofei Cai, Liang Li, Jincan Deng, Beichen Zhang, Zheng-Jun Zha, Li~Su, and Qingming Huang.
\newblock Rethinking graph neural architecture search from message-passing.
\newblock In {\em 2021 IEEE/CVF Conference on Computer Vision and Pattern Recognition (CVPR)}, pages 6653--6662, 2021.

\bibitem{relatedbotometer}
Clayton~Allen Davis, Onur Varol, Emilio Ferrara, Alessandro Flammini, and Filippo Menczer.
\newblock Botornot: A system to evaluate social bots.
\newblock In {\em Proceedings of the 25th International Conference Companion on World Wide Web}, WWW '16 Companion, page 273–274, Republic and Canton of Geneva, CHE, 2016. International World Wide Web Conferences Steering Committee.

\bibitem{yang2022botometer}
Kai-Cheng Yang, Emilio Ferrara, and Filippo Menczer.
\newblock Botometer 101: Social bot practicum for computational social scientists.
\newblock {\em Journal of Computational Social Science}, 5(2):1511--1528, 2022.

\bibitem{relatedalarfaj}
Fawaz~Khaled Alarfaj, Hassaan Ahmad, Hikmat~Ullah Khan, Abdullah~Mohammaed Alomair, Naif Almusallam, and Muzamil Ahmed.
\newblock Twitter bot detection using diverse content features and applying machine learning algorithms.
\newblock {\em Sustainability}, 15(8), 2023.

\bibitem{relatedalothali}
Eiman Alothali, Motamen Salih, Kadhim Hayawi, and Hany Alashwal.
\newblock Bot-mgat: A transfer learning model based on a multi-view graph attention network to detect social bots.
\newblock {\em Applied Sciences}, 12(16), 2022.

\bibitem{relatedsatar}
Shangbin Feng, Herun Wan, Ningnan Wang, Jundong Li, and Minnan Luo.
\newblock Satar: A self-supervised approach to twitter account representation learning and its application in bot detection.
\newblock In {\em Proceedings of the 30th ACM International Conference on Information; Knowledge Management}, CIKM ’21, page 3808–3817. ACM, 10 2021.

\bibitem{relatedilias2}
Loukas Ilias and Ioanna Roussaki.
\newblock Detecting malicious activity in twitter using deep learning techniques.
\newblock {\em Applied Soft Computing}, 107:107360, 2021.

\bibitem{relatedilias}
Loukas Ilias, Ioannis Michail~Kazelidis, and Dimitris Askounis.
\newblock Multimodal detection of bots on x (twitter) using transformers.
\newblock {\em IEEE Transactions on Information Forensics and Security}, 19:7320--7334, 2024.

\bibitem{relatedweinguyen}
Feng Wei and Uyen~Trang Nguyen.
\newblock Twitter bot detection using neural networks and linguistic embeddings.
\newblock {\em IEEE Open Journal of the Computer Society}, 4:218--230, 2023.

\bibitem{relatedmvcan}
Parisa Bazmi, Masoud Asadpour, and Azadeh Shakery.
\newblock Multi-view co-attention network for fake news detection by modeling topic-specific user and news source credibility.
\newblock {\em Information Processing \& Management}, 60(1):103146, 2023.

\bibitem{relatedbotartist}
Alexander Shevtsov, Despoina Antonakaki, Ioannis Lamprou, Polyvios Pratikakis, and Sotiris Ioannidis.
\newblock Botartist: Twitter bot detection machine learning model based on twitter suspension.
\newblock {\em arXiv preprint arXiv:2306.00037}, 2023.

\bibitem{relatedsujith}
K~Sujith, Shreya Chowdhury, Arsh Goyal, Anand~Vardhan Hegde, and Ramamoorthy Srinath.
\newblock Twitter bot detection and ranking using supervised machine learning models.
\newblock In {\em 2022 International Conference on Data Science, Agents \& Artificial Intelligence (ICDSAAI)}, volume~01, pages 1--6, 2022.

\bibitem{relatedliu}
Yuhan Liu, Zhaoxuan Tan, Heng Wang, Shangbin Feng, Qinghua Zheng, and Minnan Luo.
\newblock Botmoe: Twitter bot detection with community-aware mixtures of modal-specific experts.
\newblock In {\em Proceedings of the 46th International ACM SIGIR Conference on Research and Development in Information Retrieval}, SIGIR '23, page 485–495, New York, NY, USA, 2023. Association for Computing Machinery.

\bibitem{relatedsaxena}
Naman Saxena, Adwitiya Sinha, Tanishk Bansal, and Ankita Wadhwa.
\newblock A statistical approach for reducing misinformation propagation on twitter social media.
\newblock {\em Information Processing \& Management}, 60(4):103360, 2023.

\bibitem{relatedcaleb}
Ilias Dimitriadis, George Dialektakis, and Athena Vakali.
\newblock Caleb: A conditional adversarial learning framework to enhance bot detection.
\newblock {\em Data \& Knowledge Engineering}, 149:102245, 2024.

\bibitem{relatedyang}
Kai-Cheng Yang, Onur Varol, Pik-Mai Hui, and Filippo Menczer.
\newblock Scalable and generalizable social bot detection through data selection.
\newblock {\em Proceedings of the AAAI Conference on Artificial Intelligence}, 34(01):1096–1103, April 2020.

\bibitem{relatedquezada}
Vicente Quezada, Fabian Astudillo-Salinas, Luis Tello-Oquendo, and Paul Bernal.
\newblock Real-time bot infection detection system using dns fingerprinting and machine-learning.
\newblock {\em Computer Networks}, 228:109725, 2023.

\bibitem{relatedmiller}
Zachary Miller, Brian Dickinson, William Deitrick, Wei Hu, and Alex~Hai Wang.
\newblock Twitter spammer detection using data stream clustering.
\newblock {\em Information Sciences}, 260:64--73, 2014.

\bibitem{relatedchavoshi}
Nikan Chavoshi, Hossein Hamooni, and Abdullah Mueen.
\newblock Debot: Twitter bot detection via warped correlation.
\newblock In {\em 2016 IEEE 16th International Conference on Data Mining (ICDM)}, pages 817--822, 2016.

\bibitem{relatedbotwalk}
Amanda Minnich, Nikan Chavoshi, Danai Koutra, and Abdullah Mueen.
\newblock Botwalk: Efficient adaptive exploration of twitter bot networks.
\newblock In {\em 2017 IEEE/ACM International Conference on Advances in Social Networks Analysis and Mining (ASONAM)}, pages 467--474, 2017.

\bibitem{relatedmulbot}
L.~Mannocci, S.~Cresci, A.~Monreale, A.~Vakali, and M.~Tesconi.
\newblock Mulbot: Unsupervised bot detection based on multivariate time series.
\newblock In {\em 2022 IEEE International Conference on Big Data (Big Data)}, pages 1485--1494, Los Alamitos, CA, USA, dec 2022. IEEE Computer Society.

\bibitem{relatedwu}
Jeremy Wu, Eric Teng, and Ziyue Cao.
\newblock Twitter bot detection through unsupervised machine learning.
\newblock In {\em 2022 IEEE International Conference on Big Data (Big Data)}, pages 5833--5839, 2022.

\bibitem{relatedkoggalahewa}
Darshika Koggalahewa, Yue Xu, and Ernest Foo.
\newblock An unsupervised method for social network spammer detection based on user information interests.
\newblock {\em Journal of Big Data}, 9(1):Article number: 7, January 2022.

\bibitem{relatedlopes}
Daniele A.~G. Lopes, Marcelo~A. Marotta, Marcelo Ladeira, and João J.~C. Gondim.
\newblock Botnet detection based on network flow analysis using inverse statistics.
\newblock In {\em 2022 17th Iberian Conference on Information Systems and Technologies (CISTI)}, pages 1--6, 2022.

\bibitem{relatedalhosseini}
Seyed Ali~Alhosseini, Raad Bin~Tareaf, Pejman Najafi, and Christoph Meinel.
\newblock Detect me if you can: Spam bot detection using inductive representation learning.
\newblock In {\em Companion Proceedings of The 2019 World Wide Web Conference}, WWW '19, page 148–153, New York, NY, USA, 2019. Association for Computing Machinery.

\bibitem{relatedheterofeng}
Shangbin Feng, Zhaoxuan Tan, Rui Li, and Minnan Luo.
\newblock Heterogeneity-aware twitter bot detection with relational graph transformers.
\newblock In {\em AAAI Conference on Artificial Intelligence}, 2021.

\bibitem{relatedkusen}
Ema Kušen and Mark Strembeck.
\newblock You talkin’ to me? exploring human/bot communication patterns during riot events.
\newblock {\em Information Processing \& Management}, 57(1):102126, 2020.

\bibitem{relatedbui}
Thi Bui and Katerina Potika.
\newblock Twitter bot detection using social network analysis.
\newblock In {\em 2022 Fourth International Conference on Transdisciplinary AI (TransAI)}, pages 87--88, 2022.

\bibitem{relateddehghan}
Ali Dehghan, Krzysztof Siuta, Andrzej Skorupka, Anuja Dubey, Anthony Betlen, Derek Miller, Wei Xu, Bogdan Kamiński, and Paweł Prałat.
\newblock Detecting bots in social-networks using node and structural embeddings.
\newblock {\em Journal of Big Data}, 10(1):119, 2023.

\bibitem{relatedbot2vec}
Phu Pham, Loan~T.T. Nguyen, Bay Vo, and Unil Yun.
\newblock Bot2vec: A general approach of intra-community oriented representation learning for bot detection in different types of social networks.
\newblock {\em Information Systems}, 103:101771, 2022.

\bibitem{relatednoekhah}
Shirin Noekhah, Naomie binti Salim, and Nor~Hawaniah Zakaria.
\newblock Opinion spam detection: Using multi-iterative graph-based model.
\newblock {\em Information Processing \& Management}, 57(1):102140, 2020.

\bibitem{relatedhofa}
Sen Ye, Zhaoxuan Tan, Zhenyu Lei, Ruijie He, Hongrui Wang, Qinghua Zheng, and Minnan Luo.
\newblock Hofa: Twitter bot detection with homophily-oriented augmentation and frequency adaptive attention.
\newblock {\em arXiv preprint arXiv:2306.12870}, 2023.

\bibitem{relatedsimilcatch}
Nour El-Mawass, Paul Honeine, and Laurent Vercouter.
\newblock Similcatch: Enhanced social spammers detection on twitter using markov random fields.
\newblock {\em Information Processing \& Management}, 57(6):102317, 2020.

\bibitem{relatedzhou}
Kaixiong Zhou, Xiao Huang, Qingquan Song, Rui Chen, and Xia Hu.
\newblock Auto-gnn: Neural architecture search of graph neural networks.
\newblock {\em Frontiers in Big Data}, 5, 2022.

\bibitem{relatedgraphnas}
Yang Gao, Hong Yang, Peng Zhang, Chuan Zhou, and Yue Hu.
\newblock Graph neural architecture search.
\newblock In {\em Proceedings of the Twenty-Ninth International Joint Conference on Artificial Intelligence}, IJCAI'20, 2021.

\bibitem{relatedsnag}
Huan Zhao, Lanning Wei, and Quanming Yao.
\newblock Simplifying architecture search for graph neural network.
\newblock {\em ArXiv}, abs/2008.11652, 2020.

\bibitem{relatednunes}
Matheus Nunes and Gisele~L. Pappa.
\newblock {\em Intelligent Systems: 9th Brazilian Conference, BRACIS 2020, Rio Grande, Brazil, October 20–23, 2020, Proceedings, Part I}.
\newblock Springer International Publishing, 2020.

\bibitem{relatedmetagnas}
YuFei Li, Jia Wu, and TianJin Deng.
\newblock Meta-gnas: Meta-reinforcement learning for graph neural architecture search.
\newblock {\em Engineering Applications of Artificial Intelligence}, 123:106300, 2023.

\bibitem{relatedpeng}
Wei Peng, Xiaopeng Hong, Haoyu Chen, and Guoying Zhao.
\newblock Learning graph convolutional network for skeleton-based human action recognition by neural searching.
\newblock {\em Proceedings of the AAAI Conference on Artificial Intelligence}, 34(03):2669--2676, 4 2020.

\bibitem{relatedjiang}
S.~Jiang and P.~Balaprakash.
\newblock Graph neural network architecture search for molecular property prediction.
\newblock In {\em 2020 IEEE International Conference on Big Data (Big Data)}, pages 1346--1353, Los Alamitos, CA, USA, dec 2020. IEEE Computer Society.

\bibitem{relatedfastenas}
Yameng Peng, Andy Song, Vic Ciesielski, Haytham Fayek, and Xiaojun Chang.
\newblock Fast evolutionary neural architecture search by contrastive predictor with linear regions.
\newblock In {\em Proceedings of the Genetic and Evolutionary Computation Conference}, GECCO '23, page 1257–1266, New York, NY, USA, 2023. Association for Computing Machinery.

\bibitem{relatedshang}
Ronghua Shang, Songling Zhu, Hangcheng Liu, Teng Ma, Weitong Zhang, Jie Feng, Licheng Jiao, and Rustam Stolkin.
\newblock Evolutionary architecture search via adaptive parameter control and gene potential contribution.
\newblock {\em Swarm and Evolutionary Computation}, 82:101354, 2023.

\bibitem{relatedgea}
Vasco Lopes, Miguel Santos, Bruno Degardin, and Luís~A. Alexandre.
\newblock Guided evolutionary neural architecture search with efficient performance estimation.
\newblock {\em Neurocomputing}, page 127509, 2024.

\bibitem{relatedsane}
Huan ZHAO, Quanming YAO, and Weiwei TU.
\newblock Search to aggregate neighborhood for graph neural network.
\newblock In {\em 2021 IEEE 37th International Conference on Data Engineering (ICDE)}, pages 552--563, 2021.

\bibitem{relatedli}
Yanxi Li, Zean Wen, Yunhe Wang, and Chang Xu.
\newblock One-shot graph neural architecture search with dynamic search space.
\newblock {\em Proceedings of the AAAI Conference on Artificial Intelligence}, 35(10):8510--8517, May 2021.

\bibitem{relatedzhao}
Jiakun Zhao, Ruifeng Zhang, Zheng Zhou, Si~Chen, Ju~Jin, and Qingfang Liu.
\newblock A neural architecture search method based on gradient descent for remaining useful life estimation.
\newblock {\em Neurocomputing}, 438:184--194, 2021.

\bibitem{KANG20131384}
Ah~Reum Kang, Jiyoung Woo, Juyong Park, and Huy~Kang Kim.
\newblock Online game bot detection based on party-play log analysis.
\newblock {\em Computers \& Mathematics with Applications}, 65(9):1384--1395, 2013.
\newblock Advanced Information Security.

\bibitem{5555837}
Shahriar Mohammadi and Hossein Abbasimehr.
\newblock A high level security mechanism for internet polls.
\newblock In {\em 2010 2nd International Conference on Signal Processing Systems}, volume~3, pages V3--101--V3--105, 2010.

\bibitem{10.5555/3600270.3602825}
Shangbin Feng, Zhaoxuan Tan, Herun Wan, Ningnan Wang, Zilong Chen, Binchi Zhang, Qinghua Zheng, Wenqian Zhang, Zhenyu Lei, Shujie Yang, Xinshun Feng, Qingyue Zhang, Hongrui Wang, Yuhan Liu, Yuyang Bai, Heng Wang, Zijian Cai, Yanbo Wang, Lijing Zheng, Zihan Ma, Jundong Li, and Minnan Luo.
\newblock Twibot-22: towards graph-based twitter bot detection.
\newblock In {\em Proceedings of the 36th International Conference on Neural Information Processing Systems}, NIPS '22, Red Hook, NY, USA, 2024. Curran Associates Inc.

\bibitem{9926818}
Loukas Ilias, Dimitris Askounis, and John Psarras.
\newblock A multimodal approach for dementia detection from spontaneous speech with tensor fusion layer.
\newblock In {\em 2022 IEEE-EMBS International Conference on Biomedical and Health Informatics (BHI)}, pages 1--5, 2022.

\bibitem{ILIAS2023110834}
Loukas Ilias and Dimitris Askounis.
\newblock Context-aware attention layers coupled with optimal transport domain adaptation and multimodal fusion methods for recognizing dementia from spontaneous speech.
\newblock {\em Knowledge-Based Systems}, 277:110834, 2023.

\bibitem{dartsad}
Michail Chatzianastasis, Loukas Ilias, Dimitris Askounis, and Michalis Vazirgiannis.
\newblock Neural architecture search with multimodal fusion methods for diagnosing dementia.
\newblock In {\em ICASSP 2023 - 2023 IEEE International Conference on Acoustics, Speech and Signal Processing (ICASSP)}, pages 1--5, 2023.

\end{thebibliography}

\end{document}